%% file: arxiv.tex
\documentclass[10pt,twocolumn,letterpaper]{article}

\usepackage{iccv}
\usepackage{times}
\usepackage{epsfig}
\usepackage{graphicx}
\usepackage{amsmath}
\usepackage{amssymb}
\usepackage{mathrsfs}
\usepackage{caption}
\usepackage{graphbox}
\usepackage[dvipsnames]{xcolor}
\usepackage[accsupp]{axessibility}
\usepackage{booktabs}
\usepackage{marvosym}

\usepackage[pagebackref=true,breaklinks=true,letterpaper=true,colorlinks,bookmarks=false]{hyperref}

\usepackage[capitalize]{cleveref}
\Crefname{section}{Section}{Sections}
\crefname{section}{Sec.}{Secs.}
\Crefname{align}{Equation}{Equations}
\crefname{align}{Eq.}{Eqs.}
\Crefname{equation}{Equation}{Equations}
\crefname{equation}{Eq.}{Eqs.}
\Crefname{figure}{Figure}{Figures}
\crefname{figure}{Fig.}{Figs.}
\Crefname{table}{Table}{Tables}
\crefname{table}{Tab.}{Tabs.}

\iccvfinalcopy % *** Uncomment this line for the final submission

\def\iccvPaperID{5828} % *** Enter the ICCV Paper ID here

\ificcvfinal\fi
\begin{document}

\title{StableVideo: Text-driven Consistency-aware Diffusion Video Editing}

\author{
Wenhao Chai$^{1}$ \thanks{} \quad
Xun Guo$^{2}$\textsuperscript{\Letter} \quad
Gaoang Wang$^{1}$ \quad
Yan Lu$^{2}$\\
[2mm]
$^1$~Zhejiang University \quad $^2$~Microsoft Research Asia\\
[2mm]
{\tt\small \{wenhaochai.19, gaoangwang\}@intl.zju.edu.cn, \{xunguo, yanlu\}@microsoft.com}
}
\twocolumn[{
	\maketitle
	\vspace{-3em}
	\renewcommand\twocolumn[1][]{#1}
    \begin{center}
    \centering
    \includegraphics[width=0.99\textwidth]{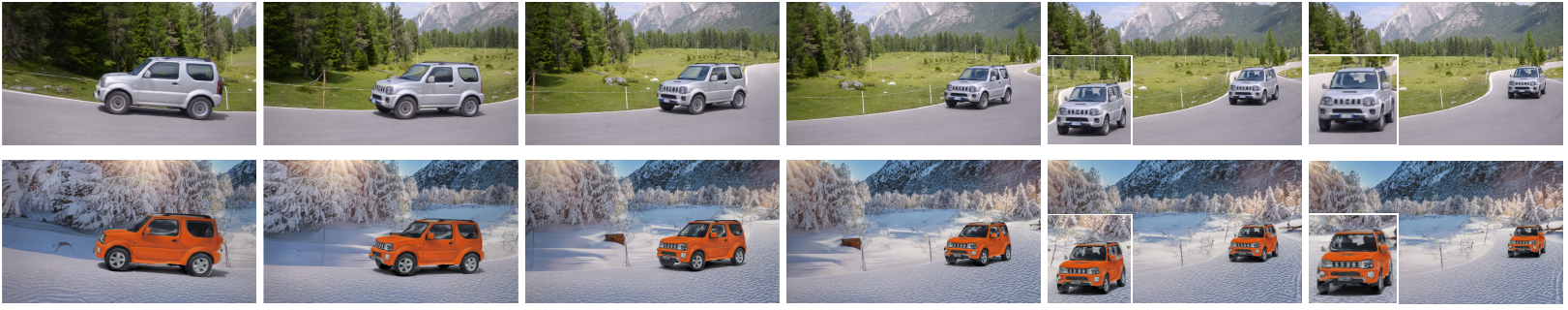}
    \captionof{figure}{
    \textbf{Editing results of StableVideo.} The input video (top row) contains long range motion and viewpoint changing. Our approach performs stable editing according to the prompt of ``\textit{Orange SUV in sunny snow winter}'' on foreground and background. In the edited video (bottom row), the ``\textit{orange SUV}'' maintains high geometric and temporal consistency, although the viewpoints keep changing.}
    \label{fig:teaser}
\end{center}
}]

{
  \renewcommand{\thefootnote}%
    {\fnsymbol{footnote}}
    \footnotetext[1]{The work was done when the author was with MSRA as an intern.}
}

\input{tex/abs.tex}
\input{tex/intro.tex}
\begin{figure*}[t!]
    \centering
    \includegraphics[width=0.99\textwidth]{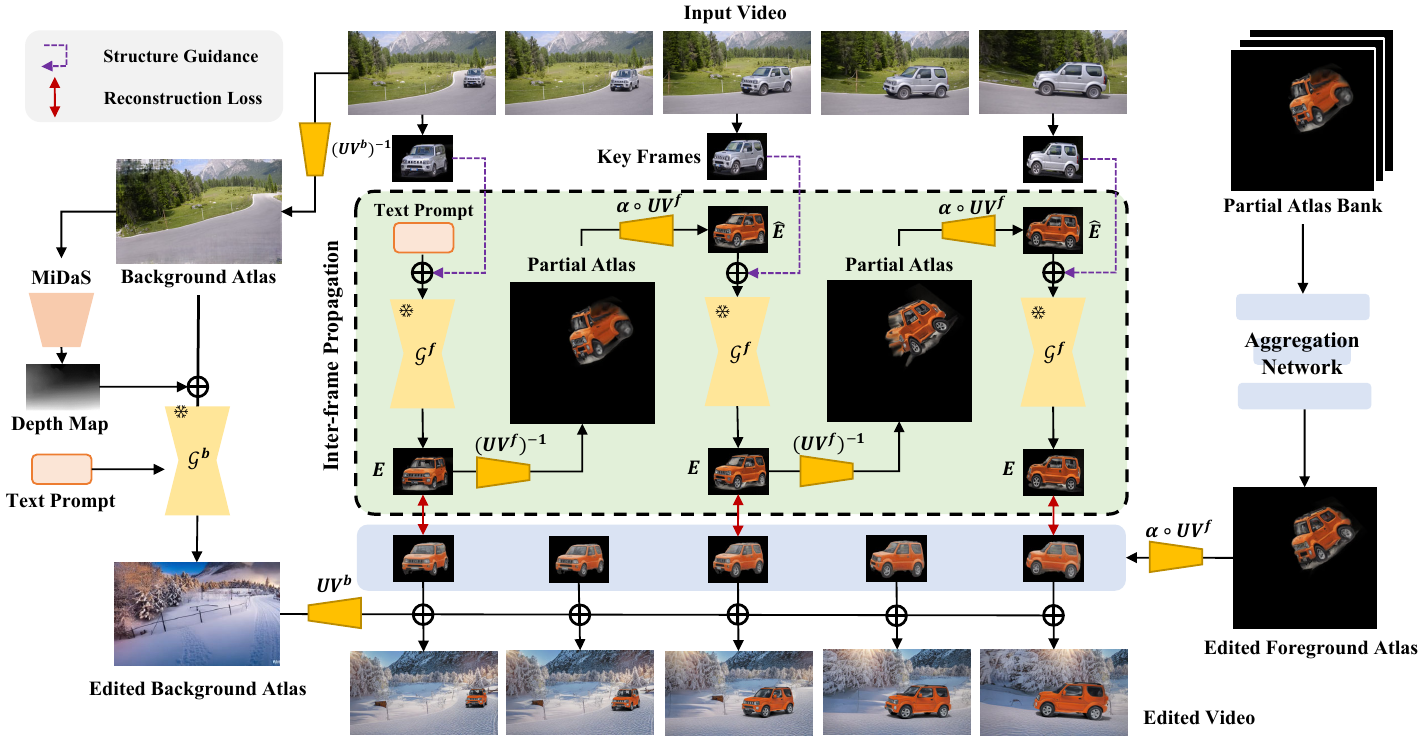}
    \caption{\textbf{Framework of the proposed StableVideo.} The input video is first fed into NLA~\cite{lu2020layered} to generate foreground and background atlases using the pre-trained model. $\mathcal{G}_b$ is the diffusion model used to edit background atlas, and $\mathcal{G}_f$ is used to edit the foreground key frames. Note that $\mathcal{G}_b$ and $\mathcal{G}_f$ share the same weights, but accept different conditions. We employ depth information, extracted by MiDaS~\cite{Ranftl2022midas}, for $\mathcal{G}_b$ to maintain the consistency between the foreground motion and the environment, while structure guidance is used for $\mathcal{G}_f$ to keep geometric consistency between the new generated foreground and the old one. After being edited, the foreground and background are blended together to reconstruct the edited frames.}
    \label{fig:framework}
\end{figure*}
\input{tex/survey.tex}
\input{tex/method.tex}
\input{tex/exp}
\input{tex/conclusion.tex}
\newpage

{\small
\bibliographystyle{ieee_fullname}
\bibliography{ref}
}

\newpage
\quad
\newpage

\input{tex/supp}

\end{document}

%% file: tex/abs.tex
\begin{abstract}
Diffusion-based methods can generate realistic images and videos, but they struggle to edit existing objects in a video while preserving their appearance over time. This prevents diffusion models from being applied to natural video editing in practical scenarios. In this paper, we tackle this problem by introducing temporal dependency to existing text-driven diffusion models, which allows them to generate consistent appearance for the edited objects. Specifically, we develop a novel inter-frame propagation mechanism for diffusion video editing, which leverages the concept of layered representations to propagate the appearance information from one frame to the next. We then build up a text-driven video editing framework based on this mechanism, namely StableVideo, which can achieve consistency-aware video editing. Extensive experiments demonstrate the strong editing capability of our approach. Compared with state-of-the-art video editing methods, our approach shows superior qualitative and quantitative results. Our code is available at \href{https://github.com/rese1f/StableVideo}{this https URL}.
\end{abstract}

%% file: tex/intro.tex
\section{Introduction}

%video diffusion
%\cite{ho2022imagenv, yu2022magvit, ho2022vdm, singer2022makeavideo}

%image diffusion

%image editing
%\cite{brooks2022instructpix2pix, parmar2023pixzero}

%video editing
%\cite{lu2020layered, bar2022text2live, wu2022tune, hertz2022prompt, meng2021sdedit}

%Challenges of video editing
Recent years have witnessed significant progress in extensive computer vision tasks taken by deep learning. Nevertheless, natural video editing, which aims at manipulating the appearance of target objects and scenes, still faces two essential challenges that are deterministic to the editing quality: the \textit{generator} equipped with rich prior knowledge that consistently produces high-fidelity edited contents adhering faithfully to the original geometry of the target objects, and the \textit{propagator} that disseminates the edited contents throughout the entire video while keeping highly temporal consistency.

%Advantages and shortages of diffusion and atlas
The flourish of text-driven generative diffusion models pre-trained on large-scale image and language data~\cite{radford2021learning, ho2022imagenv, yu2022magvit, ho2022vdm, singer2022makeavideo, chai2022deep} provides impressive generation quality. Several diffusion-based methods achieve good performance in image editing~\cite{brooks2022instructpix2pix, parmar2023pixzero}, but few methods have tried to apply diffusion models in video editing, since it is challenging to modify existing objects while preserving their appearance over the entire video~\cite{wu2022tune, hertz2022prompt, meng2021sdedit}. Dreamix~\cite{molad2023dreamix} proposes a solution to generate consistent video according to input image/video and prompts. However, it focuses more on generating smooth motions, \textit{e.g.}, pose and camera movements, rather than maintaining geometric consistency of the objects across time. Moreover, such video diffusion models often suffer from huge computing complexity which is not friendly for practical applications.\

Neural layered atlas (NLA)~\cite{lu2020layered,low2022minimal} tries to tackle the temporal continuity problem by decomposing the video into a set of atlas layers, each of which describes one target object to be edited. For each atlas layer, the positions of the video are mapped into the corresponding 2D positions in it, so that semantically correspondent pixels over the whole video can be represented by the same atlas position. Instead of frame-by-frame editing, NLA edits atlas layers to ensure that the modifications can be precisely mapped back to video frames for temporal smoothness. Text2LIVE~\cite{bar2022text2live} provides a text-driven appearance manipulation solution of adding additional edit layers on atlases, in which a specific generator for the edit layers is trained. Although it achieves good results with strict structure preserved, it is not able to apply thorough editing. Moreover, the specifically trained generator also limits the richness of the generated contents.

%motivation
This brings up the question: \textit{Could text-driven diffusion video editing achieve high temporal consistency}? Intuitively, employing text-driven diffusion models to edit the atlases corresponding to the target objects could reach such goal. However, this gives rise to drawbacks rather than benefits. Being the summary of the whole video, atlases always have distorted appearance due to the viewpoint and camera movement, which are required to be specifically pre-trained and generated as in \cite{bar2022text2live}. Diffusion models may fail in generating satisfied atlas pixels in many cases, so that the corresponding edited frames will also be contaminated. To answer the question, we present two concepts for utilizing diffusion models in video editing. Firstly, instead of editing the atlases directly, we propose to update the atlases via editing key video frames. Secondly, we introduce temporal dependency constraints for diffusion models to generate objects with consistent appearance across time. 

%We present two main ideas to enable existing diffusion models to reach this goal. Firstly, we believe that the better way for employing diffusion as generator is to perform editing on frames then aggregate the edited contents to update the corresponding atlases. Secondly, we need to develop a consistency-aware constrains on existing diffusion models to enhance their controllability, thereby generating consistent geometry on key frames.

%proposed method
Based on analysis above, we present a novel diffusion video editing approach, StableVideo, to perform consistency-aware video editing. In specific, we propose two effective technologies for this purpose. Firstly, to edit the objects with consistent appearance, we design an inter-frame propagation mechanism on top of the existing diffusion model~\cite{zhang2023adding}, which can generate new objects with coherent geometry across time. Secondly, to achieve temporal consistency by leveraging NLA, we design an aggregation network to generate the edited atlases from the key frames. We then build up a text-driven diffusion-based framework, which provides high-quality natural video editing. We conduct extensive qualitative and quantitative experiments to demonstrate the capability of our approach. Compared with state-of-the-art methods, our approach achieves superior results with much lower complexity.

%contributions
In summary, we present the following contributions:
 \begin{itemize}
     \item To our best knowledge, we are the first to solve the consistency problem of diffusion video editing by considering the concept of layered atlas approaches, which provides an efficient and effective way for this topic.
     \item We present a new video editing framework which can manipulate the appearance of the objects with high geometry and appearance consistency across time. Our method can be easily applied to other text-driven diffusion models.  
     \item We conduct extensive experiments on a variety of natural videos, which shows superior editing performance compared with state-of-the-art methods.
 \end{itemize}

%% file: tex/survey.tex
\section{Related Work}

\subsection{Diffusion for Image Editing} 
The editing of natural images is an important task in the field of computer vision that has been widely studied. Prior to the emergence of diffusion models~\cite{songdenoising,ho2020denoising}, many GAN-based approaches~\cite{goodfellow2020generative, gal2022stylegan, park2019semantic, patashnik2021styleclip, wang2018high} have achieved good results. There are also some works focus on low-level editing~\cite{ye2022perceiving,chen2022snowformer,yang2022aim}. The advent of diffusion models has made it possible to achieve even higher quality and more diverse edited contents. SDEdit~\cite{meng2021sdedit} adds noise and corruptions to an input image and uses diffusion models to reverse the process for image editing, while suffering from the loss of fidelity. Prompt-to-Prompt~\cite{hertz2022prompt} and Plug-and-Play~\cite{tumanyan2022plug} perform semantic editing by mixing activations from original and target text prompts. InstructPix2Pix~\cite{brooks2022instructpix2pix} applies semantic editing at test time and personalizes the model through finetuning and optimization to learn a special token describing the content. UniTune~\cite{valevski2022unitune} and Imagic~\cite{kawar2022imagic} finetune on a single image for better editability while maintaining good fidelity. There are also some works exploring the controllability~\cite{zhang2023adding,huang2023composer,saharia2022palette,li2023gligen,cao2023difffashion,cao2023image} and personalization~\cite{gal2022image,ruiz2022dreambooth, nikankin2022sinfusion} of diffusion-based generation. Our proposed video editing method leverages existing image editing methods that can preserve the structures, such as \cite{zhang2023adding, parmar2023pixzero,tumanyan2022plug} and \cite{mou2023t2i}.

\subsection{Diffusion for Video Editing}
Compared to image editing, video editing is more challenging for diffusion-based methods for geometric and temporal consistency. Tune-a-Video~\cite{wu2022tune} inflates a text-to-image model for video editing. However, since temporal correlations are not fully considered, the editing results suffer from inconsistency of geometry and motion. Dreamix~\cite{molad2023dreamix} develops a text-to-video backbone for motion editing while maintaining temporal consistency. There are also some works based on video generation like~\cite{esser2023structure,singer2022make,ho2022imagen,yu2023video,zhou2022magicvideo} and \cite{hovideo}. Unlike these approaches, our purpose is to enable diffusion models to perform appearance editing with both geometric and temporal consistency. 

\subsection{Temporal Propagation in Video Editing}
Temporal propagation plays an important role in natural video process, since it is the essential factor for temporal consistency. Some methods rely on key frames~\cite{jamrivska2019stylizing, texler2020interactive,xu2022temporally} or optical flow~\cite{ruder2016artistic} to propagate contents between frames. Another bunch of methods are to achieve consistent inter-frame editing by forming a compressed representation of a video. Omnimattes~\cite{lu2020layered, luassociating} estimate RGBA layers for target subject and scene effects for each frame independently, but cannot achieve consistent propagation of contents along temporal direction. Atlas~\cite{bar2022text2live, kasten2021layered} tackles this problem by decomposing the video into unified 2D atlas layers for each target. This approach allows contents to be applied to the global summarized 2D atlases and mapped back to the video, achieving temporal consistency with minimal effort. Inspired by the concept of atlas approach, we employ the pre-trained neural layered atlas model to solve the inconsistency problem in diffusion video editing, thereby achieving high-quality editing results with temporal coherence.

%% file: tex/method.tex
\section{Method}
\cref{fig:framework} shows the pipeline of our proposed StableVideo. We utilize NLA~\cite{kasten2021layered} as the propagator for consistent video editing. Specifically, we conduct foreground and background editing separately. For foreground editing, we adopt key frame editing to generate atlas layers with high quality and the inter-frame propagation module to ensure better geometric and temporal consistency. The edited key frames are then mapped to partial atlases and aggregated by the aggregation network to produce the edited foreground atlas. It is noteworthy that our approach can also handle more than one foreground layers. 

\subsection{Problem Formulation}
We employ the pre-trained NLA model~\cite{lu2020layered} to propagate the edited contents to ensure that the target objects and scenes can maintain homogeneous appearances and motions across the entire video. The concept of NLA is to decompose the input video into layered representations, namely foreground atlas and background atlas, which globally summarize the correlated pixels for the foreground and the background, respectively.  Three mapping networks, \textit{i.e.}, $\mathcal{M}^b(\cdot)$, $\mathcal{M}^f(\cdot)$ and $\mathcal{M}^{\alpha}(\cdot)$, are provided for this purpose. Given an input video $I$, for each frame $I_i$, we obtain the mapping relationships of the atlas in the background and the foreground with respect to the pixel coordinate system, named as $\mathrm{UV}^b(\cdot)$ and $\mathrm{UV}^f(\cdot)$, as well as the foreground opacity $\alpha_i$ on the pixel coordinate system, formulated as:
\begin{equation}
    \mathrm{UV}_i^b(\cdot)=\mathcal{M}^b(I_i), \mathrm{UV}_i^f(\cdot)=\mathcal{M}^f(I_i), \alpha_i=\mathcal{M}^{\alpha}(I_i).
\end{equation}
After that, we formulate the mappings from the atlas representation of the background $A^b$ and the foreground $A^f$, to the pixel coordinate systems of $B_i$ and $F_i$:
\begin{equation}
    B_i=\mathrm{UV}_i^b(A^b),F_i=\mathrm{UV}_i^f(A^f).
\end{equation}

Our method achieves geometrical consistent editing by fixing the mappings of $\mathrm{UV}^b$ and $\mathrm{UV}^f$, and generating the edited atlases of $A^b$ and $A^f$. We adopt a pre-trained latent diffusion model~\cite{rombach2022high} with guided conditions as our generator, namely $\mathcal{G}^b(\cdot)$ and $\mathcal{G}^f(\cdot)$. Note that we are not simply editing the foreground and the background atlases directly. We apply inter-frame propagation mechanism on the editing process of foreground atlas. More details are explained in Sec.~\ref{sec:inter-frame-propagation}. After that, the entire video $I$ can be reconstructed frame by frame as the following equation: 
\begin{equation}
    I_i = \alpha_i \circ \mathrm{UV}_i^f(\mathcal{G}^f(A^f)) + (1-\alpha_i) \circ \mathrm{UV}_i^b(\mathcal{G}^b(A^b)),
\end{equation}
where $\circ$ denotes pixel-wise product.

\begin{figure}[t]
    \centering
    \includegraphics[width=0.99\linewidth]{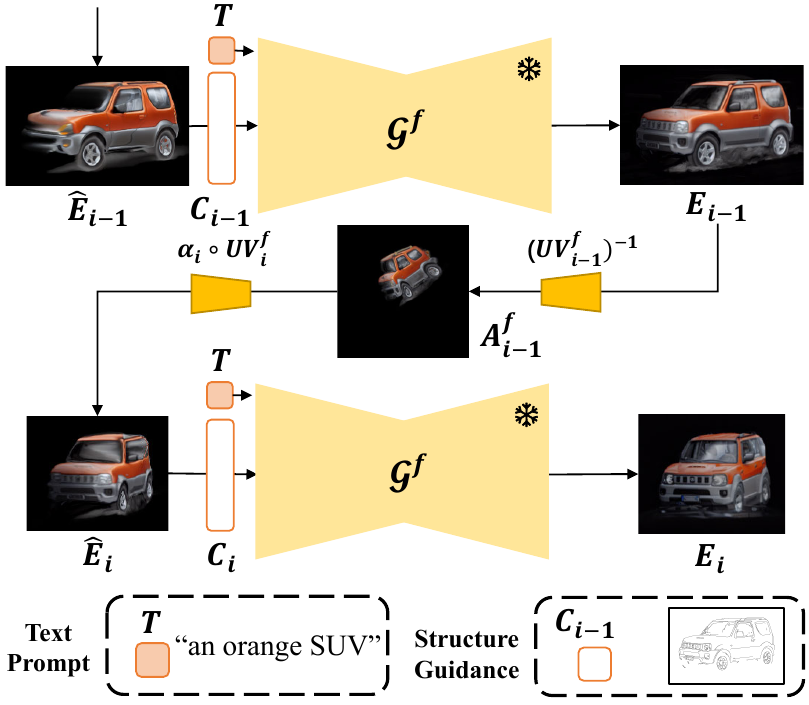}
    \caption{\textbf{Inter-frame propagation for foreground editing.} We use two edited key frames, $E_{i-1}$ and $E_i$, to illustrate the process more clearly. The structure guidance and the text prompt is added into the denoising UNet via the concatenation and cross-attention mechanism respectively.}
    \label{fig:propagate}
\end{figure}

\begin{figure}[t]
    \centering
    \includegraphics[width=0.99\linewidth]{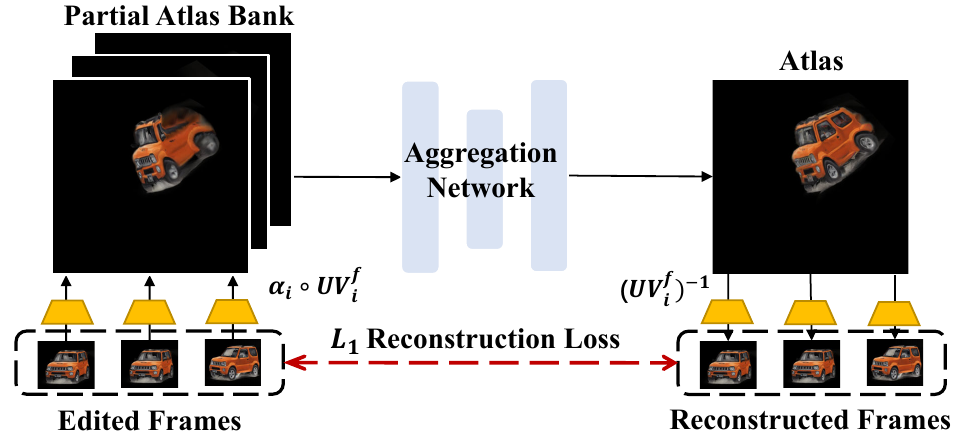}
    \caption{\textbf{Training process of aggregation network.} We employ a simple 3D network to aggregate the partial atlases generated from the edited key frames. For each training iteration, the sampled key frames are edited and mapped to partial atlases by $UV_i^f$. The partial atlases are then fed into the Aggregation Network to generate the edit atlas, which is mapped back by $(UV_i^f)^{-1}$ to generate the reconstructed frames. A $L_1$ loss is used to guarantee the aggregation consistency between the edited key frames and the reconstructed frames.}
    \label{fig:aggregate}
\end{figure}

\begin{figure*}[t]
    \centering
    \includegraphics[width=0.99\linewidth]{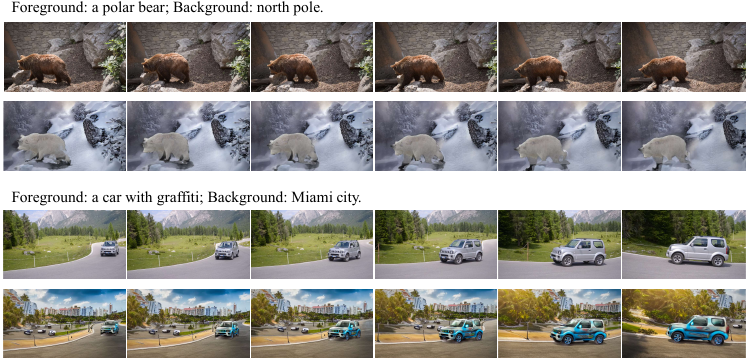}
    \caption{\textbf{Compositing editing.} We demonstrate two editing examples with non-rigid and rigid foreground objects. Our approach can well preserve the geometry of ``bear'' and ``car'' across frames.}
    \label{fig:composite_editing}
\end{figure*}

\subsection{Inter-frame Propagation}
\label{sec:inter-frame-propagation}
In this section, we further elaborate on how inter-frame propagation mechanism helps consistent foreground editing. One of the major challenges for diffusion models is to generate video contents with temporal consistency. Existing state-of-the-art text-driven diffusion methods~\cite{zhang2023adding, mou2023t2i} can maintain the similar geometry between the target objects and the generated ones for image editing by adding structure conditions. However, the situation is different for videos. Generating temporally consistent geometry needs to handle some uncertain changes across time, \textit{e.g.}, motion and deformation, which can not be supported by them. We tackle this problem by introducing a conditional denoising process to enable the diffusion models to consider both the structure of the current frame and the appearance information from previous frame, thereby sequentially generating new objects with geometric consistency across time. In specific, we employ canny edge as structure guidance, which is also adopted by existing diffusion methods~\cite{zhang2023adding}. Another important question is how to propagate the information of one object across frames to achieve consistent appearance. With the help of NLA, we can transfer the appearance features of the overlapping parts of previous frame to the next frame. Inspired by SDEdit~\cite{meng2021sdedit} and ILVR~\cite{choi2021ilvr}, we further use a process of adding noise and denoising to obtain a more complete output. We illustrate this generation process in ~\cref{fig:propagate}.  The inter-frame propagation method is only applied on the foreground objects.

Specifically, we first select $N$ foreground key frames from the original video $I$, ensuring that 1) there is significant overlap between the adjacent frames, and 2) these key frames capture the appearance of all faces of the object. Given a generator $\mathcal{G}^f(\cdot)$ and a text prompt $T$, we edit the first frame $F_0$ in pixel coordinate system with its structure condition $C_0$ as the extra guidance:
\begin{equation}
    E_0 = \mathcal{G}^f(T,C_0),
\end{equation}
where $E_0$ represents the editing result. Then for the remaining key frames, we propagate the editing result from the previous key frame $E_{i-1}$ to obtain the one of the current key frame $E_i$. To be specific, we map $E_{i-1}$ from the pixel coordinate to atlas $A^E_{i-1}$ and then map it back to the pixel coordinate in the current frame $i$ as $\hat{E}_i$ with multiplying the opacity $\alpha_i$, formulated by:
\begin{equation}
    \begin{aligned}
         A^f_{i-1} &= (\mathrm{UV}^f_{i-1})^{-1}(E_{i-1}),\\
        \hat{E}_i &= \alpha_i \circ \mathrm{UV}^f_{i}(A^f_{i-1}).
    \end{aligned}
\end{equation}
It is noteworthy that the entire video shares the same foreground atlas $A^f$ for each target object, where the subscript $i$ represents an incomplete partial atlas in frame $i$. We then use the atlas to propagate the pixel values from previous frame to their corresponding positions in the current frame to obtain an incomplete partial appearance $\hat{E}_i$. 

Given the partial appearance $\hat{E}_i$, we first encode it with VQ-VAE~\cite{van2017neural} to get the latent representation $\hat{Z}_i$, and then add noise to it with Variance Preserving Stochastic Differential Equation (VP-SDE), formulated as:
\begin{equation}
    \hat{Z}_i(t_0) = \alpha(t_0)\hat{Z}_t(0)+\sigma(t_0)\mathbf{z},\quad \mathbf{z}\sim \mathcal{N}(0,\mathbf{I})
\end{equation}
where $\sigma(t_0)$ and $\alpha(t_0)$ are two scalar functions that satisfy $\alpha^2(t)+\sigma^2(t)=1, \forall t \in (0,1]$, and  $t_0 \in [0,1] $ is a hyper parameter of the noise strength. Then we apply denoising process $\hat{Z}_i(t_0)$ under the condition guidance from both text prompt $T$ and structure guidance $C_i$ to get the latent representation $Z_i$. Finally, we decode the latent representation to $Z_i$ propagate editing result $E_i$. Our experiments further demonstrate that this mechanism can achieve good propagation results without training or fine-tuning the model.

\begin{figure*}[t]
    \centering
    \includegraphics[width=\linewidth]{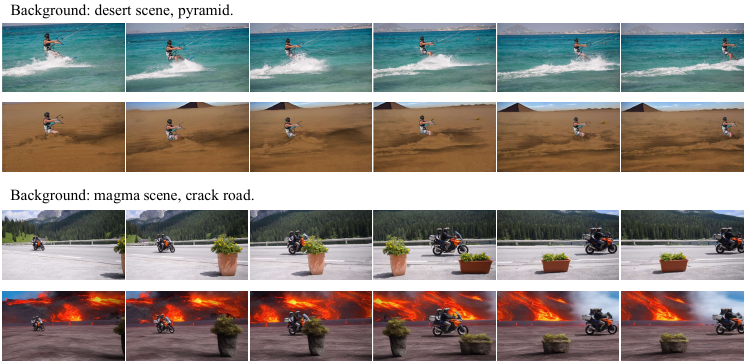}
    \caption{\textbf{Background replacement.} Since our approach can effectively maintain the geometry of the foreground, it can perform background replacement while maintaining geometric consistency of depth and temporal continuity of perspective.}
    \label{fig:background_editing}
    \vspace{-0.5cm}
\end{figure*}
\subsection{Aggregation Network}
Different from \cite{bar2022text2live} and \cite{lu2020layered}, our approach edits video frames rather than atlases, which have the chance to achieve more information of different viewpoints. This brings two advantages. Firstly, the geometries and pixels from different viewpoints provide more details of the target objects, allowing the diffusion model to generate the edited content with higher fidelity. Secondly, this alleviates the risk of failure editing due to the potential wrong mapping from the atlas to the video frames. We then aggregate the edited key frames by using a simple yet effective two-layer 2D convolution network with skip connection as shown in Fig.~\ref{fig:aggregate}. Our goal is to guarantee that the aggregated atlas is highly aligned with the original one, in terms of locations, so that appearance edit will not affect the geometric consistency and the temporal continuity. Reconstruction loss, $\mathcal{L}_{rec}$, between the edited and reconstructed key frames is employed in the training process as:
 \begin{align}
     \label{eq:aggnetloss}
     \mathcal{L}_{rec}=\sum_{i=1}^N||E_i-\mathrm{UV}_i^f(A^f)||_1.
 \end{align}
where $N$ is the number of key frames.

%% file: tex/exp.tex
\section{Experiments}

\subsection{Experimental Settings}
In practice, we implement our approach over Stable Diffusion~\cite{rombach2022high}. Despite there are several image-based methods~\cite{parmar2023pixzero,tumanyan2022plug,mou2023t2i} can perform structure-preserving editing, we choose the canny condition branch from \cite{zhang2023adding} as the structure guidance for the proposed inter-frame propagation in our method. We apply our method on several videos from DAVIS~\cite{pont20172017}, with each video containing a \textit{moving} object in 50 $\sim$ 70 frames. The image resolution is set to 768 $\times$ 432, and the resolution of foreground atlas is set to 2000 $\times$ 2000.  We employ DDIM~\cite{song2020denoising} sampler with 20 steps. In this case,  our method requires only $\sim$10 GB GPU memory and takes $\sim$30 seconds for each video in a single NVIDIA A40 GPU.

\subsection{Editing Results} 

% \begin{figure}[h]
%     \centering
%     \includegraphics[width=0.99\linewidth]{fig/foreground_editing.pdf}
%     \caption{\textbf{Foreground attribute edit.} We demonstrate the editing effect for three attributes, \textit{i.e.}, color, species and material. Our method can maintain the shape of the foreground unchanged throughout the video. }
%     \label{fig:foreground_editing}
% \end{figure}

\begin{figure*}[h]
    \centering
    \includegraphics[width=0.99\linewidth]{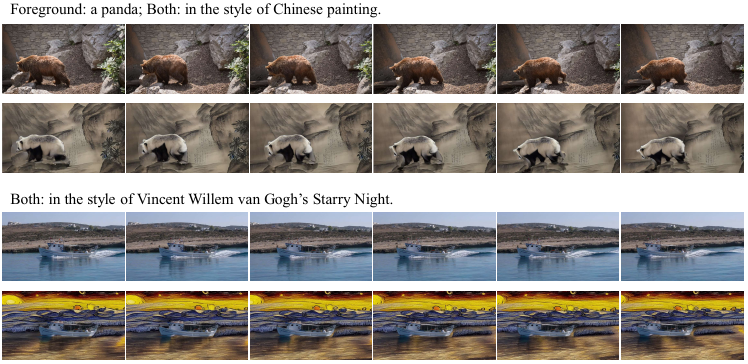}
    \caption{\textbf{Style transfer.} Our approach achieves diverse video style transfer while ensuring high temporal consistency.}
    \label{fig:style_editing}
\end{figure*}

We tested our method for various editing types among several videos. Here we demonstrate several scenarios: 

\noindent\textbf{Compositing editing.} Our method can edit foreground and background separately to achieve high-quality and semantically matched editing results as shown in Fig.~\ref{fig:composite_editing}.

% \noindent\textbf{Foreground attribute edit.} Our method is able to edit various attributes of foreground objects in videos using text prompts, and remain the shape of the foreground unchanged throughout the video as shown in Fig.~\ref{fig:foreground_editing}. The editable attributes include color, species, shapes and texture.

\noindent\textbf{Background replacement.} Due to the separate editing capabilities enabled by NLA, as well as the diverse and high-quality editing brought by the diffusion model, our method achieves high-quality video background replacement while maintaining geometric consistency of depth and temporal continuity of perspective, as shown in Fig.~\ref{fig:background_editing}.

\noindent\textbf{Style transfer.} Our method accomplishes a wide range of style transfers, while simultaneously ensuring temporal consistency, as shown in Fig.~\ref{fig:style_editing}.

\subsection{Comparison to Prior Arts}
In this section, we compare our editing results with state-of-the-art atlas-based method Text2LIVE~\cite{bar2022text2live} and diffusion-based method Tune-A-Video~\cite{wu2022tune}.

\begin{figure}[t]
    \centering
    \includegraphics[width=\linewidth]{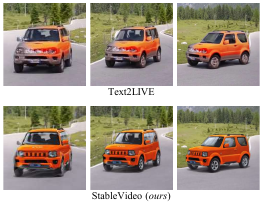}
    \caption{\textbf{Comparison to Text2LIVE.} Foreground prompt: an orange SUV. Our method achieves more holistic editing results on the foreground.}
    \label{fig:compare_text2live}
    \vspace{-10pt}
\end{figure}

\noindent\textbf{Comparison to Text2LIVE~\cite{bar2022text2live}.} As shown in Fig.~\ref{fig:compare_text2live}, given the text prompt \textit{``an orange SUV''}, Text2LIVE shows incomplete editing, while our method demonstrates a much more holistic result. It is because our method employs key frame editing, while Text2LIVE creates a new layer to edit the atlas somehow directly. The potential failure editing in atlas will lead to the error propagation to the entire video. In addition, compared to Text2LIVE, we can achieve higher quality and richer content with faster inference speed.

\begin{figure}[t]
    \centering
    \includegraphics[width=\linewidth]{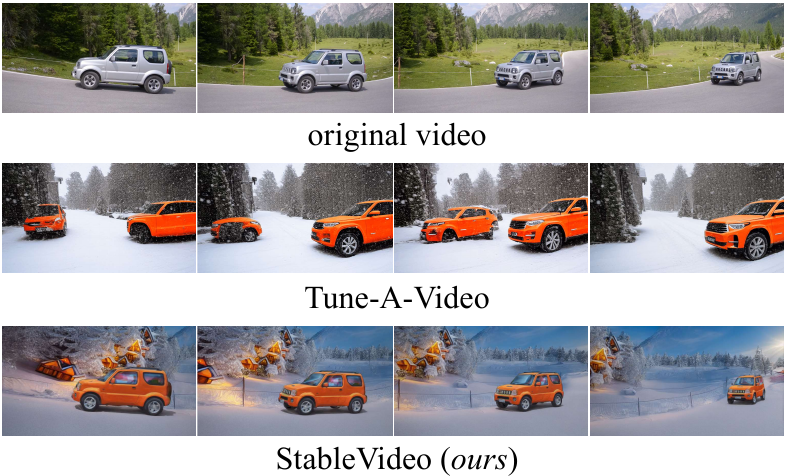}
    \caption{\textbf{Comparison to Tune-A-Video.} Prompt: an orange SUV in sunny snowy winter, cabins. Our method achieves much more consistent editing results.}
    \label{fig:compare_tuneavideo}
\end{figure}

\begin{figure}[htb]
    \centering
    \includegraphics[width=\linewidth]{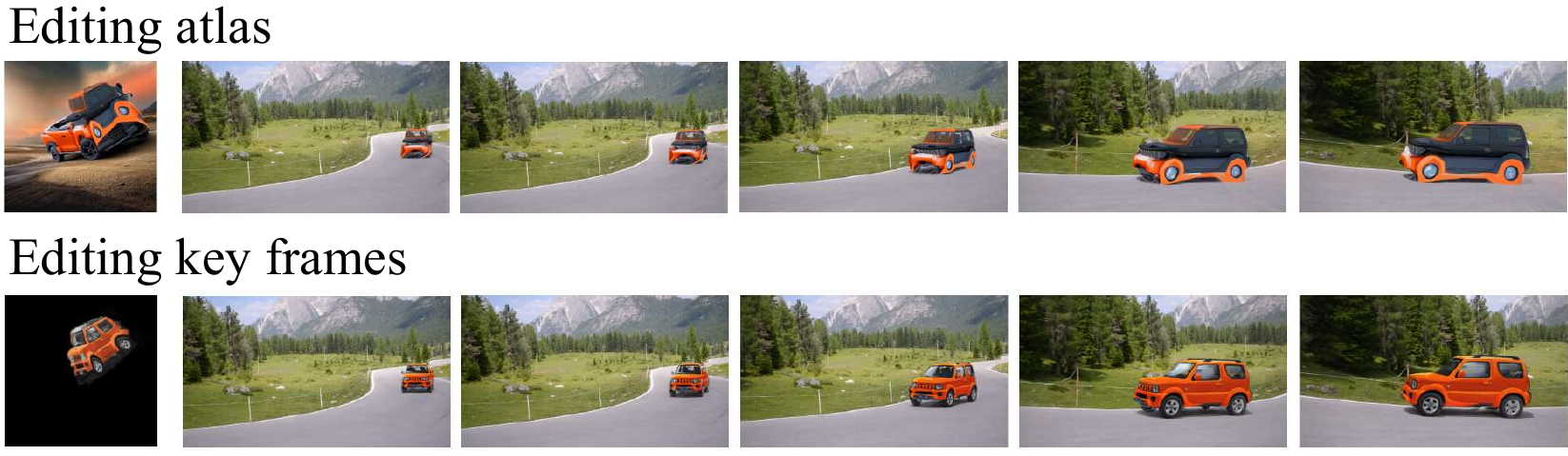}
    \caption{\textbf{Ablation study on directly editing the atlas.} The deformation in atlas affects the diffusion models.}
    \label{fig:r31}
\end{figure}

\noindent\textbf{Comparison to Tune-A-Video~\cite{wu2022tune}.} As shown in Fig.~\ref{fig:compare_tuneavideo}, vanilla video diffusion models like Tune-A-Video often fail in video editing. While it adeptly captures the semantic information of text prompts, it struggles to preserve consistency in video layout and object geometry.

\noindent\textbf{Consistency analysis.}
To the best of our knowledge, there is currently no widely accepted metric for evaluating the geometric and temporal consistency of videos. In this paper, we employ motion consistency of dense optical flow  and deviation consistency of the edited video frames for this metric. For motion consistency, we employ the Farneback algorithm in OpenCV to calculate the average L2 distance of dense optical flow between the edited and original videos. We select the "car-turn" video from DAVIS and apply the same prompt to all methods. The experiment is repeated several times to obtain the average number. Our method shows better stability than Tune-A-Video as shown in Tab.~\ref{tab:optical_flow}. For deviation consistency, we conduct experiments on 1) CLIP score: target text faithfulness 2) LPIPS-P: deviation from the original video frames and 3) LPIPS-T: deviation between adjacent frames, as shown in~\cref{tab:r24}. Our method achieves comparable CLIP score and much lower deviations, which shows the effectiveness and stability.

\begin{figure}[t]
    \centering
    \includegraphics[width=\linewidth]{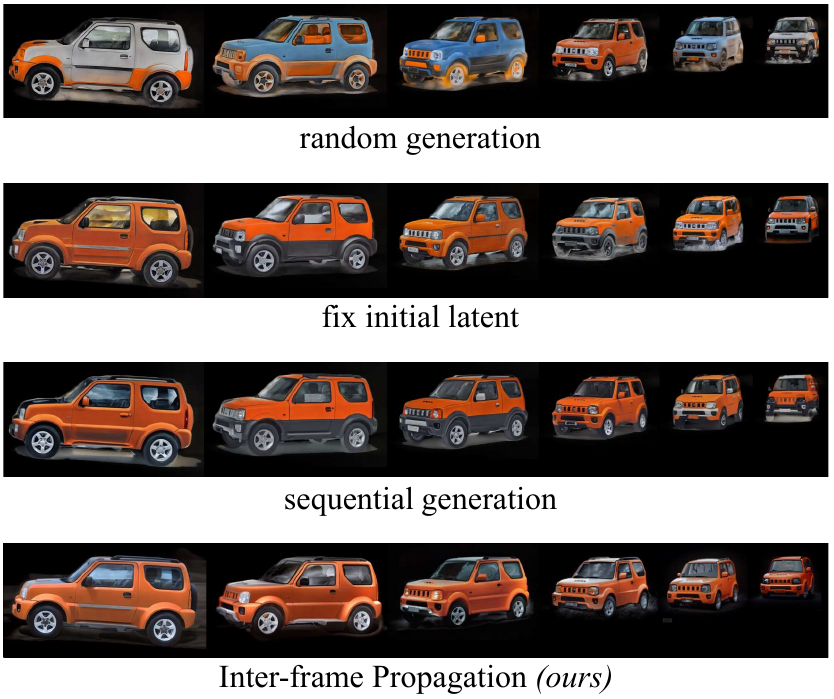}
    \caption{\textbf{Ablation study on inter-frame propagation module.} Foreground prompt: an orange SUV. Our method achieves excellent consistency in key frame appearance.}
    \label{fig:ablation_propagete}
\end{figure}

\begin{figure}[t]
    \centering
    \includegraphics[width=\linewidth]{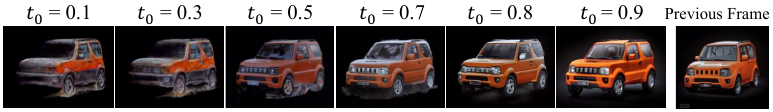} 
    \caption{\textbf{Binary search on hyper parameter  $t_0$ in inter-frame propagation module.} The appearance of previous frame is shown on the right side. The sequence on the left shows the generated results of the current frame as a function of $t_0$. The value of $t_0$ around 0.8 is a good sweet spot.}
    \label{fig:ablation_t}
    \vspace{-10pt}
\end{figure}

\subsection{Ablation Study}

To verify the necessity of the key frame editing, we apply editing in atlas layer directly for the foreground as a simple baseline. The atlas might not be so deformed for human perception, but it significantly affects the diffusion models. Fig.~\ref{fig:r31} shows an example, where obvious deformation and inconsistency exist.

We also conduct extensive ablation study on inter-frame propagation module. The objective of this module is to maintain the geometry of the foreground when editing key frames. Firstly, we consider four different settings for editing key frames as shown in Fig.~\ref{fig:ablation_propagete}.

\noindent\textbf{Random generation.} Each key frame only shares the same text prompt with the others. In this case, there are significant differences among the generated key frames.

\noindent\textbf{Fix initial latent.} Unlike starting from random noise every time we edit, we start generating each key frame from the same latent noise and share the text prompt. In this case, there is higher similarity in the content generated for each frame, but the consistency is still not satisfactory.

\noindent\textbf{Sequential generation.} Furthermore, we concatenate the latent noise between frames. Specifically, we apply image-to-image translation between frames. This method still cannot guarantee consistency since the appearances of the objects between the two frames do not match.

\noindent\textbf{Inter-frame propagation (ours).} Our final approach is to employ partial atlas to geometrically align the appearances between two frames, followed by a process of adding noise and then apply denoising process. 

In addition, We also conducted experiments on the selection of the hyper parameter $t_0$ in this module as shown in Fig.~\ref{fig:ablation_t}. We observe that as $t_0$ gradually increases, the generated results become more realistic but gradually lose their match with the appearance of the previous frame. Using binary search, we find out that a reasonable trade-off between fidelity and realism for $t_0$ lies around $0.8$.

\begin{table}[t]
    \centering
    \resizebox{0.9\linewidth}{!}{
    \begin{tabular}{c|cc}
         \toprule
         Method &  Foreground Editing & Composite Editing\\
         \midrule
         Text2LIVE~\cite{bar2022text2live} & 1.99 & 7.39\\
         Tune-A-Video~\cite{wu2022tune} & 4.63 & 12.74\\
         \midrule
         StableVideo (\textit{ours}) & 3.34 & 11.48\\ 
         \bottomrule
    \end{tabular}
    }
    \caption{\textbf{Consistency of dense optical flow.} Original video: car-turn. Foreground prompt (column 1): an orange suv. Composite prompt (column 2): a car driving in winter.}
    \label{tab:optical_flow}
\end{table}

\begin{table}[t]
    \centering
    \resizebox{0.9\linewidth}{!}{
    \begin{tabular}{c|ccc}
         \toprule
         Method & CLIP~($\uparrow$) & LPIPS-P~($\downarrow$) & LPIPS-T~($\downarrow$) \\
         \midrule
         Tune-A-Video~\cite{wu2022tune} & 0.2787 & 0.6346 & 0.1851\\
         \midrule
         StableVideo (\textit{ours}) & 0.2713 & 0.1613 & 0.0386 \\ 
         \bottomrule
    \end{tabular}
    }
    \caption{\textbf{Quantitative results.} Original video: car-turn. Prompt: an orange suv in the winter.}
\label{tab:r24}
\end{table}

%% file: tex/conclusion.tex
\begin{figure}[t]
    \centering
    \includegraphics[width=\linewidth]{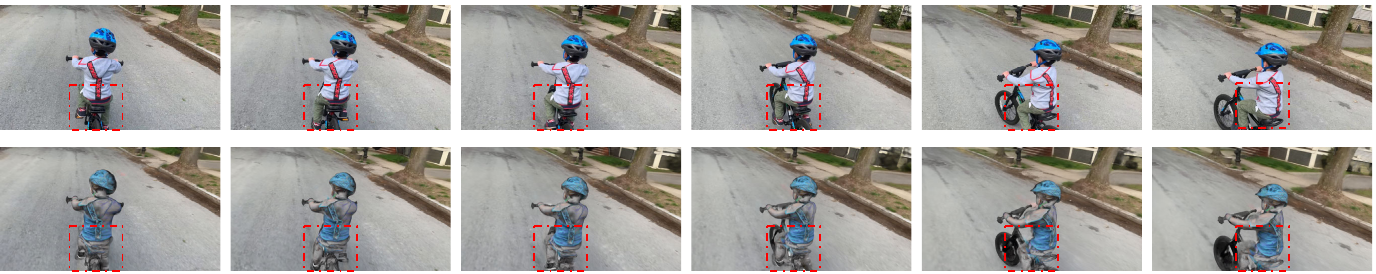}
    \caption{\textbf{Failure case.} Videos with non-rigid deformation may lead to failure editing, since the movement of the object is more difficult to be well captured in this case.}
    \label{fig:r23}
    \vspace{-10pt}
\end{figure}
\section{Limitations and Future Works}
% As a preliminary attempt towards the temporal consistency problem of diffusion models, our inter-frame propagation module is using the pre-trained diffusion model without involving any finetuning, which may limits the performance of our approach. It may better to make the generation more effective might be optimizing the diffusion model with the objective of aligning the generated objects to the reconstructed ones. Moreover, since our approach highly depend on the performance of NLA, failure cases may occur when there are complex motions and occlusions.

Firstly, our method is constrained by NLA. Learning atlas layers may fail for non-rigid objects with significant structural deformation as shown in Fig.~\ref{fig:r23}. While we can mitigate this by dividing long videos into short clips where the objects can be considered to be rigid, it is still not feasible to address every single case. Secondly, our method is constrained by the capabilities of the diffusion models, which may struggle with specific scenarios such as human or animals. Besides, it may be better to optimize the diffusion model with the objective of aligning the generated contents to the reconstructed ones.

\section{Conclusion}
We have proposed a text-driven diffusion video editing approach. To solve the consistency problem for diffusion models in foreground object editing, we propose a inter-frame propagation mechanism and an atlas aggregation network. We conducted extensive experiments and demonstrated the superior qualitative and quantitative results of our method compared to state-of-the-art approaches.

%% file: tex/supp.tex
\section*{Supplement Material}

\subsection*{A. Implementation Details}
In our experiments, we choose key frames for foreground editing by evenly sampling the input frames, \textit{i.e.}, every 20 frames. We train the aggregation network for 500 epochs with initial learning rate of 0.003 and momentum of 0.9. The network consists of two convolution layers with a ReLU in between, for which the training process is very fast. At inference stage, we conduct the training once for each edit. We set the lower and upper thresholds of Canny edges as 100 and 200 respectively, which can make the edges better represent the structure of the foreground. The numbers in Tab.~1 are the optical flow differences between the videos before and after editing (lower is better). We use \textit{cv2.calcOpticalFlowFarneback} with default parameters. More detailed setting could be found in our code that will be released soon.

\subsection*{B. Failure Cases}
Since our approach edits the key frames by using existing pre-trained diffusion models, some failure cases will occur due to the ineffective diffusion control. For example, our inter-frame propagation can well preserve the structure of the target objects across time, but cannot guarantee the quality of partial editing, as shown in \cref{fig:failure_girl}. This problem could be handled by using the masks provided by the users in practical applications, which would be our future work. As we discussed in the manuscript, NLA~\cite{lu2020layered} may fail to build the foreground atlas due to the complex motion or occlusion. In this case, our editing will also fail. However, since our approach edits directly on key frames and generates corresponding partial atlases, such failure can be alleviated.

\renewcommand{\thefigure}{A}
\begin{figure}[b]
    \centering
    \includegraphics[width=0.99\linewidth]{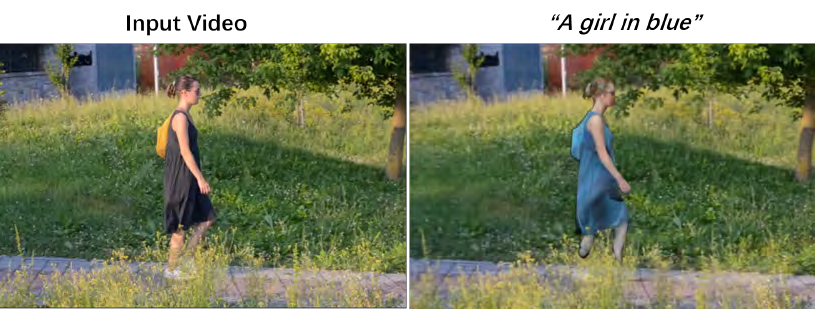} 
    \caption{An example of failure editing. Our method generates the edited contents by leveraging existing diffusion models~\cite{zhang2023adding, rombach2022high}. In the case of partial editing, \textit{e.g.}, changing the color of the skirt, the diffusion models may generate the whole person instead.}
    \label{fig:failure_girl}
\end{figure}

\renewcommand{\thefigure}{B}
\begin{figure*}[t]
    \centering    \includegraphics[width=0.99\linewidth]{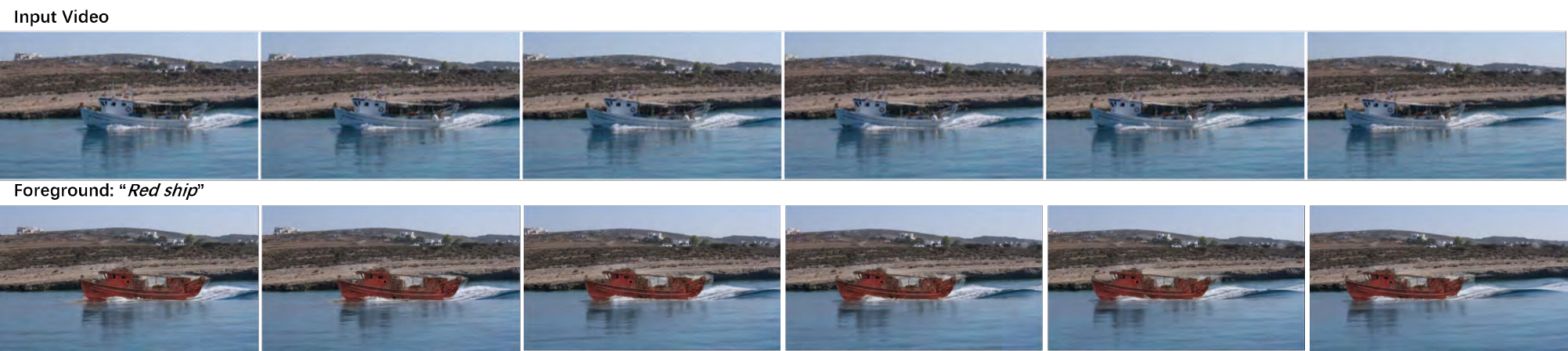}
    \caption{The editing results of foreground. The ship in this video has relatively complex geometry. Our approach can well preserve the temporal consistency. }
    \label{fig:ship_foreground}
\end{figure*}

\renewcommand{\thefigure}{C}
\begin{figure*}[t]
    \centering    \includegraphics[width=0.99\linewidth]{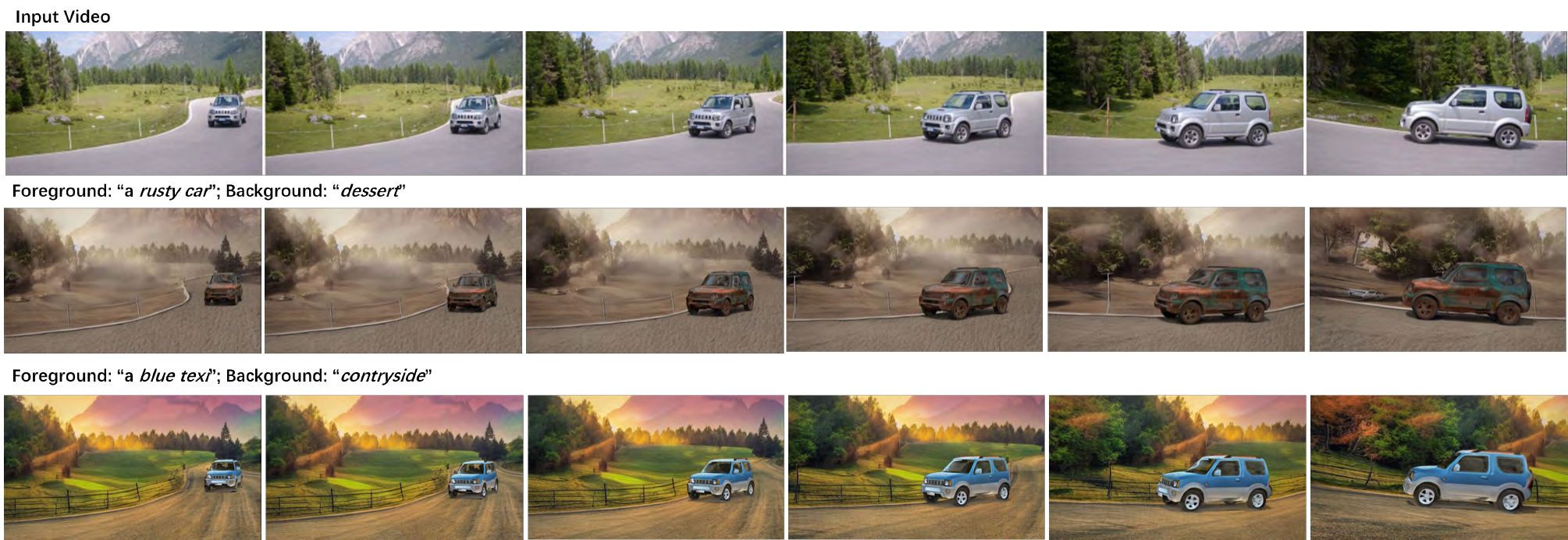}
    \caption{The results of composite editing. We separately edit the foreground and the background with semantically correlated prompts.}
    \label{fig:car_foreground}
\end{figure*}

\subsection*{C. Complexity Analysis}
Since inference  is also an essential factor for video editing, we provide the comparison of our approach to existing state-of-the-art methods, \textit{i.e.}, Tune-A-Video~\cite{wu2022tune} and Text2LIVE~\cite{bar2022text2live} as shown in \cref{tab:time}. Our approach only needs to perform lightweight training for atlas aggregation at inference stage, thereby being more efficient in pratical application compared to Text2LIVE and Tune-A-Video.

\renewcommand{\thetable}{A}
\begin{table}[t]
    \centering
    %\footnotesize
    \resizebox{\linewidth}{!}{
    \begin{tabular}{c|ccc}
    \hline
       Method & Video Training & Edit Training & Edit Inference\\
    \hline
       Text2LIVE~\cite{bar2022text2live} & $\sim$ 10 hr & $\sim$ 1 hours & $\sim$ 10 sec\\
       Tune-A-Video~\cite{wu2022tune} & $\sim$ - & 30 min & $\sim$ 4 min\\
       StableVideo (\textit{ours}) & $\sim$ 10 hr & - & $\sim$ 30 sec\\
    \hline
    \end{tabular}
    }
    \caption{The inference speed of three methods. Video Training: training once for each video. Edit Training: training once for each edit. Edit Inference: inference time. The approximated cost time is tested under the video with 768 $\times$ 432 resolution and 70 frames in a single NVIDIA A40. For StableVideo, we pick three key frames for foreground editing.}
    \label{tab:time}
\end{table}

\subsection*{D. More Editing Results}
We provide more editing results to demonstrate the effectiveness of our approach. \cref{fig:ship_foreground} shows the foreground editing for the video of "boat". We can see that the temporal consistency is well preserved. \cref{fig:car_foreground} shows the composite edit of our approach. Since the foreground and background are generated by the same diffusion model, they are highly semantically consistent. Besides, the geometry is also be well preserved across time.